\pdfoutput=1

\documentclass{article}
\usepackage{iclr2018}
\usepackage{times}
\usepackage{float}
\usepackage{hyperref}
\usepackage{url}
\usepackage{graphicx}
\usepackage{amsmath}
\usepackage{amsfonts}
\usepackage{subcaption}
\usepackage{listings}
\usepackage{lipsum}
\usepackage{microtype}
\usepackage{booktabs}
\usepackage{multirow}
\usepackage{siunitx}
\usepackage{hyphenat}

\usepackage{pifont}
\newcommand{\cmark}{\ding{51}}%
\newcommand{\xmark}{\ding{55}}%

\newcommand{\mixup}{\emph{mixup}}

\DeclareMathOperator*{\E}{\mathbb{E}}

\lstdefinestyle{mypython}{
  language=python,
  breaklines=true,
  basicstyle=\fontsize{8.5}{13}\selectfont\ttfamily,
  keywordstyle=\bfseries\color{green!40!black},
}

\iclrfinalcopy

\title{\mixup{}: Beyond Empirical Risk Minimization}

\author{%
Hongyi Zhang \\
MIT
\And
Moustapha Cisse, Yann N. Dauphin, David Lopez-Paz\thanks{Alphabetical order.} \\
FAIR
}

\begin{document}
    \maketitle
    \begin{abstract}
        Large deep neural networks are powerful, but exhibit undesirable
        behaviors such as memorization and sensitivity to adversarial examples.
        In this work, we propose \mixup{}, a simple learning principle to
        alleviate these issues. In essence, \mixup{} trains a neural network on
        convex combinations of pairs of examples and their labels.  By doing
        so, \mixup{} regularizes the neural network to favor simple linear
        behavior in-between training examples.  Our experiments on the
        ImageNet-2012, CIFAR-10, CIFAR-100, Google commands and UCI datasets
        show that \mixup{} improves the generalization of state-of-the-art
        neural network architectures.  We also find that \mixup{} reduces the
        memorization of corrupt labels, increases the robustness to adversarial
        examples, and stabilizes the training of generative adversarial
        networks.
    \end{abstract}
    \section{Introduction}

Large deep neural networks have enabled breakthroughs in fields such as
computer vision \citep{krizhevsky2012imagenet}, speech
recognition~\citep{hinton2012deep}, and reinforcement
learning~\citep{silver2016mastering}.  In most successful applications, these
neural networks share two commonalities.  First, they are trained as to
minimize their average error over the training data, a learning rule also known
as the Empirical Risk Minimization (ERM) principle \citep{vapnik98}.  Second,
the size of these state-of-the-art neural networks scales linearly with the
number of training examples.  For instance, the network of
\citet{springenberg2014striving} used $10^6$ parameters to model the $5 \cdot
10^4$ images in the CIFAR-10 dataset, the network of \citep{simonyan2014very}
used $10^8$ parameters to model the $10^6$ images in the ImageNet-2012 dataset,
and the network of \citet{chelba2013one} used $2 \cdot 10^{10}$ parameters to
model the $10^9$ words in the One Billion Word dataset. 

Strikingly, a classical result in learning theory \citep{vapnik1971uniform}
tells us that the convergence of ERM is guaranteed as long as the size of the
learning machine (e.g., the neural network) does not increase with the number
of training data. Here, the size of a learning machine is measured in terms of
its number of parameters or, relatedly, its VC-complexity
\citep{harvey2017nearly}.

This contradiction challenges the suitability of ERM to train our current
neural network models, as highlighted in recent research.  On the one hand, ERM
allows large neural networks to \emph{memorize} (instead of \emph{generalize}
from) the training data even in the presence of strong regularization, or in
classification problems where the labels are assigned at random
\citep{2016arXiv161103530Z}. On the other hand, neural networks trained with
ERM change their predictions drastically when evaluated on examples just
outside the training distribution \citep{SzegedyZSBEGF13}, also known as
\emph{adversarial examples}.  This evidence suggests that ERM is unable to
explain or provide generalization on testing distributions that differ
\emph{only slightly} from the training data. However, what is the alternative
to ERM? 

The method of choice to train on similar but different examples to the training
data is known as \emph{data augmentation} \citep{simard1998transformation},
formalized by the Vicinal Risk Minimization (VRM) principle \citep{vicinal}. In
VRM, human knowledge is required to describe a \emph{vicinity} or neighborhood
around each example in the training data. Then, additional \emph{virtual}
examples can be drawn from the vicinity distribution of the training examples
to enlarge the support of the training distribution. For instance, when
performing image classification, it is common to define the vicinity of one
image as the set of its horizontal reflections, slight rotations, and mild
scalings.  While data augmentation consistently leads to improved
generalization \citep{simard1998transformation}, the procedure is
dataset-dependent, and thus requires the use of expert knowledge.  Furthermore,
data augmentation assumes that the examples in the vicinity share the same
class, and does not model the vicinity relation across examples of different
classes.

\paragraph{Contribution} Motivated by these issues, we introduce a simple and
data-agnostic data augmentation routine, termed \mixup{}
(Section~\ref{sec:mixup}).  In a nutshell, \mixup{} constructs virtual training
examples 
\begin{align*}
  \tilde{x} &= \lambda x_i + (1 - \lambda) x_j,\qquad \text{where~} x_i, x_j \text{~are~raw~input~vectors}\\
  \tilde{y} &= \lambda y_i + (1 - \lambda) y_j,\qquad \text{where~} y_i, y_j \text{~are~one-hot~label~encodings}
\end{align*}
$(x_i, y_i)$ and $(x_j, y_j)$ are two examples drawn at random from our
training data, and $\lambda \in [0,1]$. Therefore, \mixup{} extends the
training distribution by incorporating the prior knowledge that linear
interpolations of feature vectors should lead to linear interpolations of
the associated targets.  \mixup{} can be implemented in a few lines of code,
and introduces minimal computation overhead. 

Despite its simplicity, \mixup{} allows a new state-of-the-art performance in
the CIFAR-10, CIFAR-100, and ImageNet-2012 image classification datasets
(Sections~\ref{sec:imagenet} and \ref{sec:cifar}). Furthermore, \mixup{}
increases the robustness of neural networks when learning from corrupt labels
(Section~\ref{sec:corrupt}), or facing adversarial examples
(Section~\ref{sec:adversarial}).  Finally, \mixup{} improves generalization on
speech (Sections~\ref{sec:speech}) and tabular (Section~\ref{sec:uci}) data,
and can be used to stabilize the training of GANs (Section~\ref{sec:gans}). The
source-code necessary to replicate our CIFAR-10 experiments is available at:
\begin{center}
\url{https://github.com/facebookresearch/mixup-cifar10}.
\end{center}
To understand the effects of various design choices in \mixup{}, we conduct a thorough set of ablation study experiments (Section~\ref{sec:ablation}). The results suggest that \mixup{} performs significantly better than related methods in previous work, and each of the design choices contributes to the final performance. We conclude by exploring the connections to prior work
(Section~\ref{sec:related}), as well as offering some points for discussion
(Section~\ref{sec:discussion}).

    \section{From Empirical Risk Minimization to \mixup{}}
\label{sec:mixup}

In supervised learning, we are interested in finding a function $f \in
\mathcal{F}$ that describes the relationship between a random feature vector
$X$ and a random target vector $Y$, which follow the joint distribution $P(X,
Y)$.  To this end, we first define a loss function $\ell$ that penalizes the
differences between predictions $f(x)$ and actual targets $y$, for examples
$(x, y) \sim P$.  Then, we minimize the average of the loss function $\ell$
over the data distribution $P$, also known as the \emph{expected risk}:
\begin{equation*}
    R(f) = \int \ell(f(x), y) \mathrm{d} P(x,y).
\end{equation*}
Unfortunately, the distribution $P$ is unknown in most practical situations.
Instead, we usually have access to a set of training data $\mathcal{D} =
\{(x_i, y_i)\}_{i=1}^n$, where $(x_i, y_i) \sim P$ for all $i = 1, \ldots, n$.
Using the training data $\mathcal{D}$, we may approximate $P$ by the
\emph{empirical distribution}
\begin{equation*}
    P_\delta(x, y) = \frac{1}{n} \sum_{i=1}^n \delta(x = x_i, y = y_i),
\end{equation*}
where $\delta(x = x_i, y = y_i)$ is a Dirac mass centered at $(x_i, y_i)$.
Using the empirical distribution $P_\delta$, we can now approximate the
expected risk by the \emph{empirical risk}:
\begin{equation}
    R_\delta(f) = \int \ell(f(x), y) \mathrm{d} P_\delta(x,y) = \frac{1}{n}
    \sum_{i=1}^n \ell(f(x_i), y_i). \label{eq:erm}
\end{equation}
Learning the function $f$ by minimizing \eqref{eq:erm} is known as the
Empirical Risk Minimization (ERM) principle \citep{vapnik98}.  While efficient
to compute, the empirical risk \eqref{eq:erm} monitors the behaviour of $f$
only at a finite set of $n$ examples.  When considering functions with a number
parameters comparable to $n$ (such as large neural networks), one trivial way
to minimize \eqref{eq:erm} is to memorize the training data
\citep{2016arXiv161103530Z}. Memorization, in turn, leads to the undesirable
behaviour of $f$ outside the training data \citep{SzegedyZSBEGF13}.
 
However, the na\"ive estimate $P_\delta$ is one out of many possible choices to
approximate the true distribution $P$. For instance, in the \emph{Vicinal Risk
Minimization} (VRM) principle \citep{vicinal}, the distribution $P$ is
approximated by
\begin{equation*}
	P_\nu(\tilde{x}, \tilde{y}) = \frac{1}{n} \sum_{i=1}^n \nu(\tilde{x},
\tilde{y} | x_i, y_i),
\end{equation*}
where $\nu$ is a \emph{vicinity distribution} that measures the probability of
finding the \emph{virtual} feature-target pair $(\tilde{x}, \tilde{y})$ in the
\emph{vicinity} of the training feature-target pair $(x_i, y_i)$. In
particular, \citet{vicinal} considered Gaussian vicinities $\nu(\tilde{x},
\tilde{y} | x_i, y_i) = \mathcal{N}(\tilde{x} - x_i, \sigma^2) \delta(\tilde{y}
= y_i)$, which is equivalent to augmenting the training data with additive
Gaussian noise. To learn using VRM, we sample the vicinal distribution to
construct a dataset $\mathcal{D}_\nu := \{(\tilde{x}_i,
\tilde{y}_i)\}_{i=1}^m$, and minimize the \emph{empirical vicinal risk}:
\begin{equation*}
    R_\nu(f) = \frac{1}{m}
    \sum_{i=1}^m \ell(f(\tilde{x}_i), \tilde{y}_i). \label{eq:vrm}
\end{equation*}
The contribution of this paper is to propose a generic vicinal
distribution, called \mixup{}:
\begin{equation*}
   \mu(\tilde{x}, \tilde{y} | x_i, y_i) = \frac{1}{n} \sum_j^n\E_{\lambda} \left[ \delta(\tilde{x}
   = \lambda \cdot x_i + (1-\lambda) \cdot x_j, \tilde{y} = \lambda \cdot y_i +
   (1-\lambda) \cdot y_j) \right],
\end{equation*}
where $\lambda \sim \text{Beta}(\alpha, \alpha)$, for $\alpha \in (0, \infty)$.
In a nutshell, sampling from the \mixup{} vicinal distribution produces virtual
feature-target vectors
\begin{align*}
  \tilde{x} &= \lambda x_i + (1 - \lambda) x_j,\\
  \tilde{y} &= \lambda y_i + (1 - \lambda) y_j,
\end{align*}
where $(x_i, y_i)$ and $(x_j, y_j)$ are two feature-target vectors drawn at
random from the training data, and $\lambda \in [0, 1]$. The \mixup{}
hyper-parameter $\alpha$ controls the strength of interpolation between
feature-target pairs, recovering the ERM principle as $\alpha \to 0$.

\begin{figure}
\begin{subfigure}[b]{0.65\textwidth}
\begin{lstlisting}[style=mypython]
# y1, y2 should be one-hot vectors
for (x1, y1), (x2, y2) in zip(loader1, loader2):
    lam = numpy.random.beta(alpha, alpha)
    x = Variable(lam * x1 + (1. - lam) * x2)
    y = Variable(lam * y1 + (1. - lam) * y2)
    optimizer.zero_grad()
    loss(net(x), y).backward()
    optimizer.step()
\end{lstlisting}
    \caption{One epoch of \mixup{} training in PyTorch.}
    \label{fig:mixup:code}
\end{subfigure}
\hfill
\begin{subfigure}[b]{0.35\textwidth}
    \includegraphics[width=\textwidth]{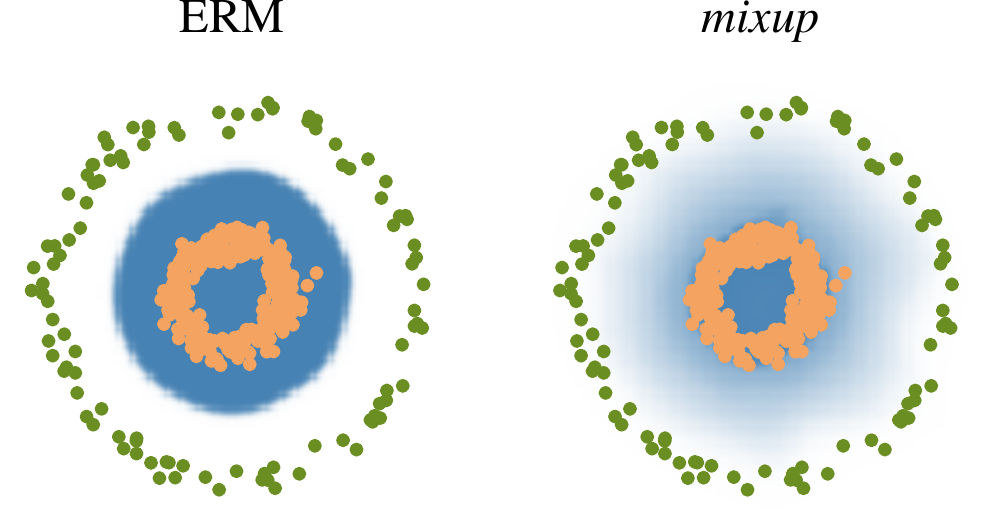}
    \caption{Effect of \mixup{} ($\alpha=1$) on a toy problem. Green: Class 0. Orange: Class 1. Blue shading indicates $p(y=1|x)$.}
    \label{fig:mixup:toy}
\end{subfigure}
\label{fig:mixup}
\vspace{-15pt}
\caption{Illustration of \mixup{}, which converges to ERM as $\alpha \to 0$.}
\end{figure}

The implementation of \mixup{} training is straightforward, and introduces a
minimal computation overhead. Figure~\ref{fig:mixup:code} shows the few lines
of code necessary to implement \mixup{} training in PyTorch. Finally, we
mention alternative design choices. First, in preliminary experiments we find that
convex combinations of three or more examples with weights sampled from a
Dirichlet distribution does not provide further gain, but increases the computation cost
of \mixup{}. Second, our current implementation uses a single data loader to
obtain one minibatch, and then \mixup{} is applied to the same minibatch after
random shuffling. We found this strategy works equally well, while reducing I/O
requirements. Third, interpolating only between inputs with equal label did not
lead to the performance gains of \mixup{} discussed in the sequel. More empirical comparison can be found in Section \ref{sec:ablation}.

\begin{figure}
	\begin{subfigure}[t]{0.48\textwidth}
		\centering
		\includegraphics[width=0.75\textwidth]{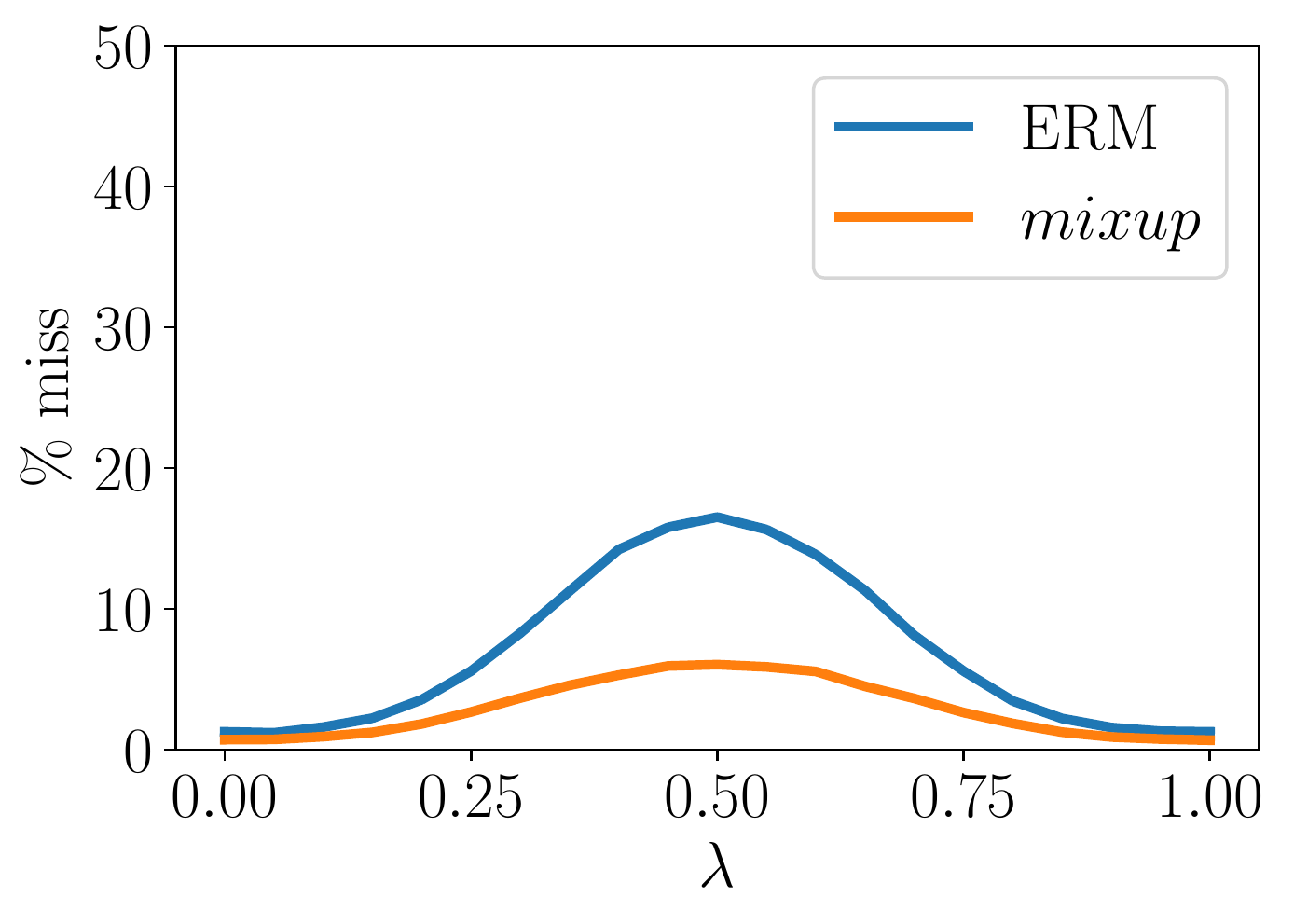}
        \caption{Prediction errors in-between training data. Evaluated
        at $x = \lambda x_i + (1-\lambda)x_j$, a prediction is counted as a
        ``miss'' if it does not belong to $\{y_i, y_j\}$. The model trained
        with \mixup{} has fewer misses.}
		\label{fig:cifar10_interp:miss}
	\end{subfigure}
	\hfill
	\begin{subfigure}[t]{0.48\textwidth}
		\centering
		\includegraphics[width=0.75\textwidth]{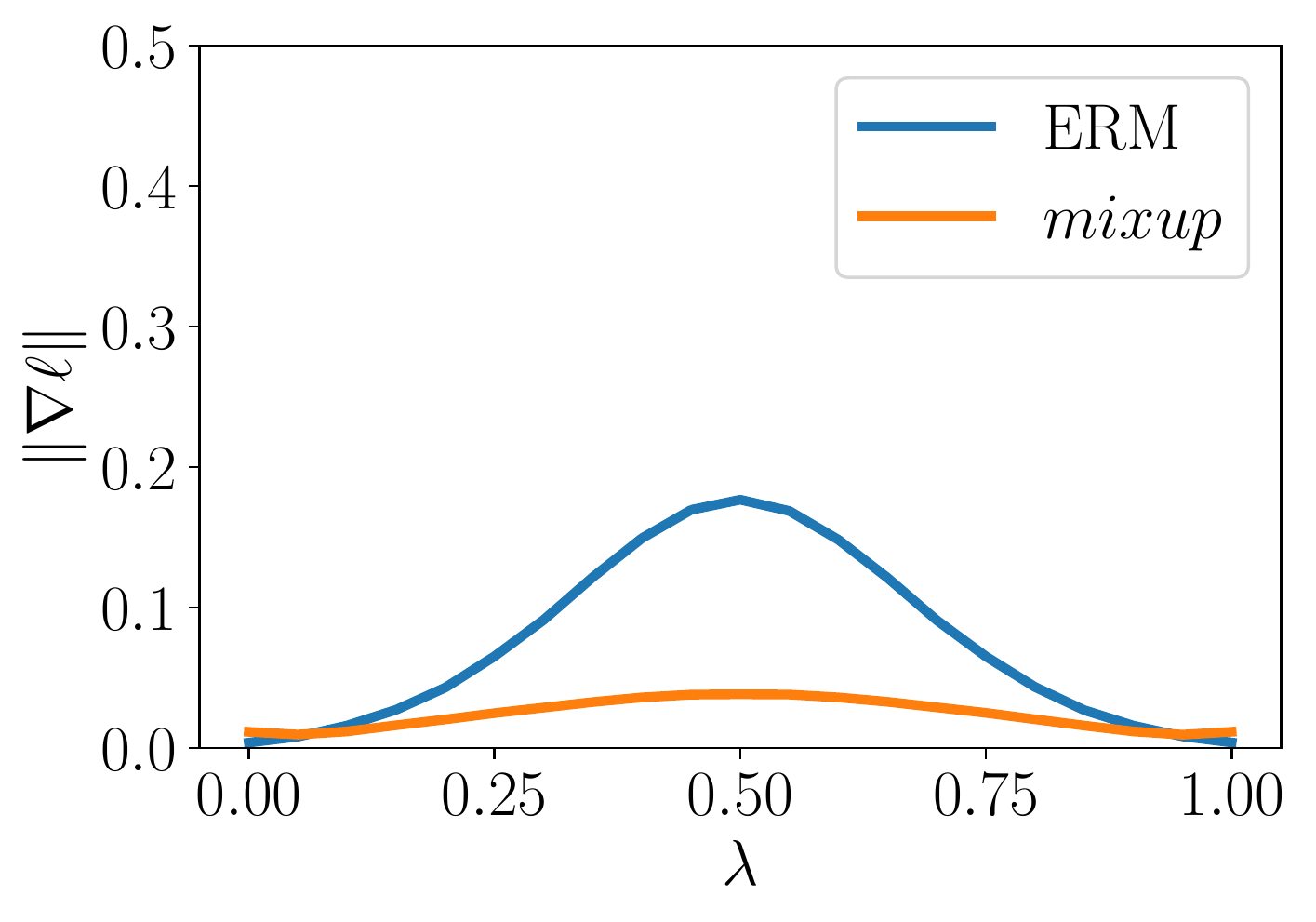}
        \caption{Norm of the gradients of the model w.r.t. input in-between training data,
        evaluated at $x = \lambda x_i + (1-\lambda)x_j$. The model trained with
        \mixup{} has smaller gradient norms.}
		\label{fig:cifar10_interp:norm}
	\end{subfigure}
	\caption{\mixup{} leads to more robust model behaviors in-between the training data.}
	\label{fig:cifar10_interp}
\end{figure}

\paragraph{What is \mixup{} doing?} The \mixup{} vicinal distribution can be
understood as a form of data augmentation that encourages the model $f$ to
behave linearly in-between training examples. We argue that this linear
behaviour reduces the amount of undesirable oscillations when predicting
outside the training examples. Also, linearity is a good inductive bias from
the perspective of Occam's razor, since it is one of the simplest possible
behaviors. Figure~\ref{fig:mixup:toy} shows that \mixup{} leads to decision
boundaries that transition linearly from class to class, providing a smoother
estimate of uncertainty. Figure~\ref{fig:cifar10_interp} illustrate the average
behaviors of two neural network models trained on the CIFAR-10 dataset using
ERM and \mixup{}.  Both models have the same architecture, are trained with the
same procedure, and are evaluated at the same points in-between randomly
sampled training data.  The model trained with \mixup{} is more stable in terms
of model predictions and gradient norms in-between training samples.

    \section{Experiments}

\subsection{ImageNet classification}
\label{sec:imagenet}
We evaluate \mixup{} on the ImageNet-2012 classification dataset
\citep{ILSVRC15}. This dataset contains 1.3 million training images and 50,000
validation images, from a total of 1,000 classes. For training, we follow
standard data augmentation practices: scale and aspect ratio distortions,
random crops, and horizontal flips \citep{goyal2017accurate}. During
evaluation, only the $224\times 224$ central crop of each image is tested. We
use \mixup{} and ERM to train several state-of-the-art ImageNet-2012
classification models, and report both top-1 and top-5 error rates in
Table~\ref{table:imagenet_result}.

\begin{table}
	\centering
	\begin{tabular}[b]{ ll rrr }
		\toprule
		{Model} & {Method} & {Epochs} & {Top-1 Error} &  {Top-5 Error} \\
        \midrule
		{ResNet-50} & ERM \citep{goyal2017accurate} & $90$ & $23.5$ & {-} \\
		& \mixup{} $\alpha=0.2$ & $90$ & {$\bf 23.3$} &  {$\bf 6.6$}\\
		\cmidrule(lr){2-2}
		{ResNet-101} & ERM \citep{goyal2017accurate} & $90$ & $22.1$ & {-} \\
		& \mixup{} $\alpha=0.2$ & $90$ & {$\bf 21.5$} &  {$\bf 5.6$}\\
		\cmidrule(lr){2-2}
		{ResNeXt-101 32*4d} & ERM \citep{xie2016aggregated} & $100$ & $21.2$ & {-} \\
		& ERM & $90$ & $21.2$ & $5.6$ \\
		& \mixup{} $\alpha=0.4$ & $90$ & {$\bf 20.7$} &  {$\bf 5.3$}\\
		\cmidrule(lr){2-2}
        ResNeXt-101 64*4d  & ERM \citep{xie2016aggregated} & $100$ & $20.4$ & $5.3 $ \\
        & \mixup{} $\alpha=0.4$ & $90$ & {$\bf 19.8$} & {$\bf 4.9 $} \\
		\midrule
		{ResNet-50} & ERM & $200$ & $23.6$ & $7.0$ \\
		& \mixup{} $\alpha=0.2$ & $200$ & {$\bf 22.1$} &  {$\bf 6.1$}\\
		\cmidrule(lr){2-2}
		{ResNet-101} & ERM & $200$ & $22.0$ & $6.1$ \\
		 & \mixup{} $\alpha=0.2$ & $200$ & {$\bf 20.8$} &  {$\bf 5.4$}\\
		 \cmidrule(lr){2-2}
		{ResNeXt-101 32*4d}  & ERM & $200$ & $21.3$ & $5.9$ \\
		& \mixup{} $\alpha=0.4$ & $200$ & $\bf 20.1$ &  $\bf 5.0$\\
        \bottomrule
	\end{tabular}
    \caption{Validation errors for ERM and \mixup{} on the development set of
    ImageNet-2012.}
    \label{table:imagenet_result}
\end{table}

For all the experiments in this section, we use data-parallel distributed
training in Caffe2\footnote{\url{https://caffe2.ai}} with a minibatch size of 1,024. We use the learning rate schedule
described in \citep{goyal2017accurate}. Specifically, the learning rate is
increased linearly from 0.1 to 0.4 during the first 5 epochs, and it is then
divided by 10 after 30, 60 and 80 epochs when training for 90 epochs; or after
60, 120 and 180 epochs when training for 200 epochs.

For \mixup{}, we find that $\alpha \in [0.1, 0.4]$ leads to improved
performance over ERM, whereas for large $\alpha$, \mixup{} leads to
underfitting. We also find that models with higher capacities and/or longer
training runs are the ones to benefit the most from \mixup{}. For example, when
trained for 90 epochs, the \mixup{} variants of ResNet-101 and ResNeXt-101
obtain a greater improvement (0.5\% to 0.6\%) over their ERM analogues than the
gain of smaller models such as ResNet-50 (0.2\%). When trained for 200 epochs,
the top-1 error of the \mixup{} variant of ResNet-50 is further reduced by
1.2\% compared to the 90 epoch run, whereas its ERM analogue stays the same.

\subsection{CIFAR-10 and CIFAR-100}
\label{sec:cifar}
We conduct additional image classification experiments on the CIFAR-10 and
CIFAR-100 datasets to further evaluate the generalization performance of
\mixup{}. In particular, we compare ERM and \mixup{} training for: PreAct
ResNet-18 \citep{he2016identity} as implemented in \citep{cifar-pytorch},
WideResNet-28-10 \citep{Zagoruyko2016WRN} as implemented in
\citep{wide-sergey}, and DenseNet \citep{huang2017densely} as implemented in
\citep{dense-andreas}. For DenseNet, we change the growth rate to 40 to follow
the DenseNet-BC-190 specification from \citep{huang2017densely}.  For \mixup{},
we fix $\alpha=1$, which results in interpolations $\lambda$ uniformly
distributed between zero and one. All models are trained on a single Nvidia
Tesla P100 GPU using PyTorch\footnote{\url{http://pytorch.org}} for 200 epochs on the training set with 128 examples per
minibatch, and evaluated on the test set. Learning rates start at 0.1 and are
divided by 10 after 100 and 150 epochs for all models except WideResNet. For
WideResNet, we follow \citep{Zagoruyko2016WRN} and divide the learning rate by
10 after 60, 120 and 180 epochs. Weight decay is set to $10^{-4}$. We do not
use dropout in these experiments.

\begin{figure}
	\centering
	\begin{subfigure}{0.6\textwidth}
		\centering
		\begin{tabular}[b]{ llrr }
            \toprule
			Dataset & Model & ERM &  \mixup{} \\
			\midrule
			\multirow{3}{*}{CIFAR-10} & PreAct ResNet-18 & $5.6$ &  $\bf 4.2$\\
			& WideResNet-28-10 & $3.8$ &  $\bf 2.7$\\
			& DenseNet-BC-190 & $3.7$ &  $\bf 2.7$\\
			\midrule
			\multirow{3}{*}{CIFAR-100} & PreAct ResNet-18 & $25.6$ &  $\bf 21.1$\\
			& WideResNet-28-10 & $19.4$ &  $\bf 17.5$\\
			& DenseNet-BC-190 & $19.0$ &  $\bf 16.8$\\
            \bottomrule
		\end{tabular}
        \caption{Test errors for the CIFAR experiments.}
        \label{fig:cifar_results:table}
	\end{subfigure}
	\hfill
	\begin{subfigure}{0.33\textwidth}
		\includegraphics[width=\textwidth]{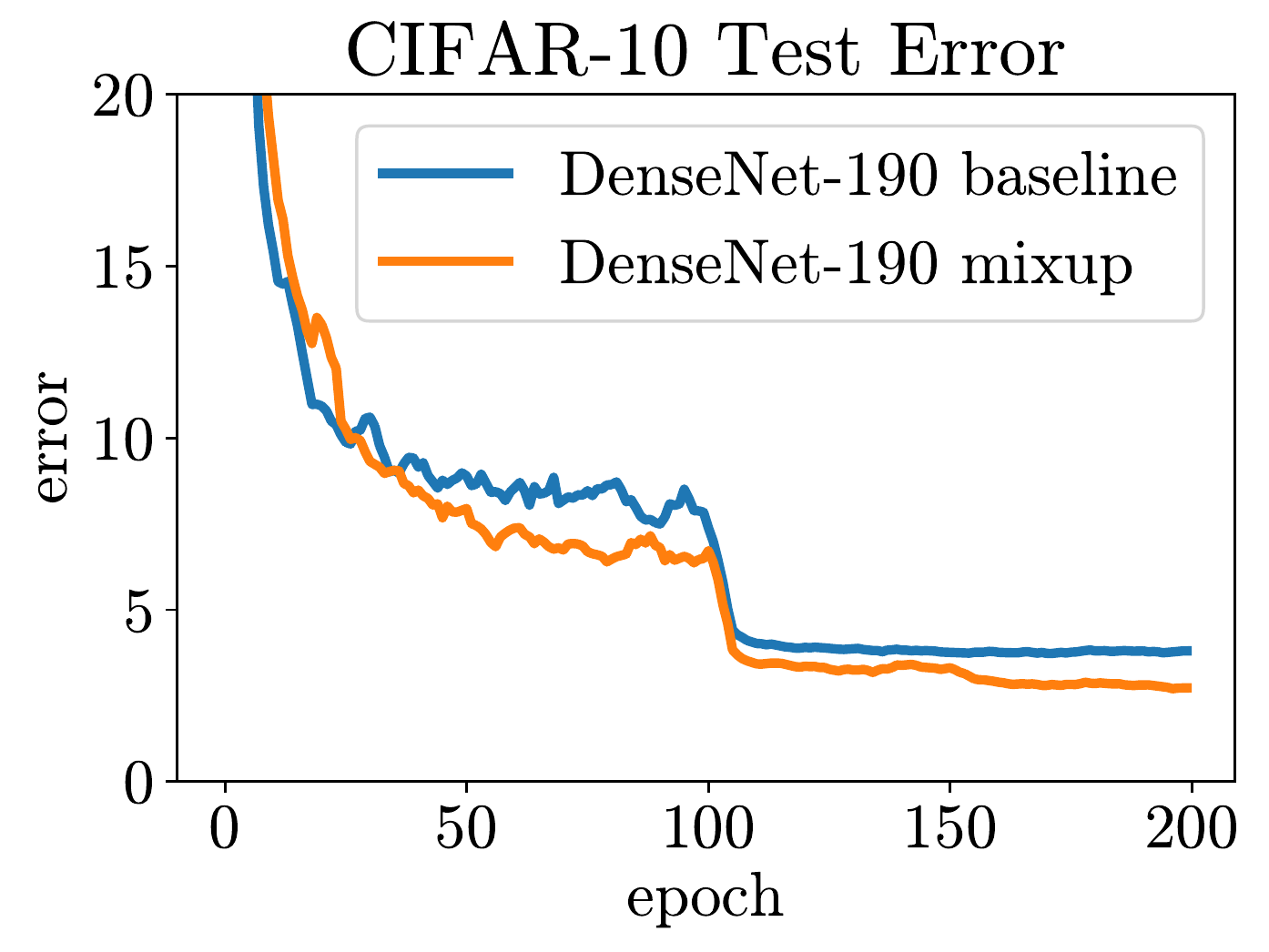}
        \caption{Test error evolution for the best ERM and \mixup{} models.}
        \label{fig:cifar_results:plot}
	\end{subfigure}
    \caption{Test errors for ERM and \mixup{} on the CIFAR experiments.}
    \label{fig:cifar_results}
\end{figure}

We summarize our results in Figure \ref{fig:cifar_results:table}. In both
CIFAR-10 and CIFAR-100 classification problems, the models trained using
\mixup{} significantly outperform their analogues trained with ERM. As seen in
Figure~\ref{fig:cifar_results:plot}, \mixup{} and ERM converge at a similar
speed to their best test errors.
Note that the DenseNet models in \citep{huang2017densely} were trained for 300
epochs with further learning rate decays scheduled at the 150 and 225 epochs,
which may explain the discrepancy the performance of DenseNet reported in
Figure \ref{fig:cifar_results:table} and the original result of
\citet{huang2017densely}.

\subsection{Speech data}
\label{sec:speech}
Next, we perform speech recognition experiments using the Google commands
dataset \citep{commands}. The dataset contains 65,000 utterances, where each
utterance is about one-second long and belongs to one out of 30 classes. The
classes correspond to voice commands such as \emph{yes, no, down, left}, as
pronounced by a few thousand different speakers. To preprocess the utterances,
we first extract normalized spectrograms from the original waveforms at a
sampling rate of 16 kHz. Next, we zero-pad the spectrograms to equalize their
sizes at $160 \times 101$. For speech data, it is reasonable to apply \mixup{}
both at the waveform and spectrogram levels. Here, we apply \mixup{} at the
spectrogram level just before feeding the data to the network.

For this experiment, we compare a LeNet \citep{lecun98} and a VGG-11
\citep{simonyan2014very} architecture, each of them composed by two
convolutional and two fully-connected layers. We train each model for 30 epochs
with minibatches of 100 examples, using Adam as the optimizer
\citep{kingma2014adam}. Training starts with a learning rate equal to
$3\times10^{-3}$ and is divided by 10 every 10 epochs.  For \mixup{}, we use a
warm-up period of five epochs where we train the network on original training
examples, since we find it speeds up initial convergence.
Table~\ref{fig:speech_results} shows that \mixup{} outperforms ERM on this
task, specially when using VGG-11, the model with larger capacity.

\begin{figure}
	\centering
		\begin{tabular}[b]{ll rr }
            \toprule
			Model & Method & Validation set & Test set\\
            \midrule
			\multirow{3}{*}{LeNet}  & ERM                     & $\bf 9.8$ & $\bf  10.3$\\
			                        & \mixup{} $(\alpha=0.1)$ &     $10.1$ &      $10.8$\\
			                        & \mixup{} $(\alpha=0.2)$ &    $10.2$ &      $11.3$\\
			\midrule
			\multirow{3}{*}{VGG-11} & ERM                     &     $5.0$ &      $4.6$\\
			                        & \mixup{} $(\alpha=0.1)$ &     $4.0$ &      $3.8$\\
			                        & \mixup{} $(\alpha=0.2)$ & $\bf 3.9$ &  $\bf 3.4$\\
            \bottomrule
		\end{tabular}
	\caption{Classification errors of ERM and \mixup{} on the Google commands dataset.}
    \label{fig:speech_results}
\end{figure}

\subsection{Memorization of corrupted labels}
\label{sec:corrupt}

Following \cite{2016arXiv161103530Z}, we evaluate the robustness of ERM and
\mixup{} models against randomly corrupted labels.  We hypothesize that
increasing the strength of \mixup{} interpolation $\alpha$ should generate
virtual examples further from the training examples, making memorization more
difficult to achieve.  In particular, it should be easier to learn
interpolations between real examples compared to memorizing interpolations
involving random labels. We adapt an open-source implementation
\citep{random-labels} to generate three CIFAR-10 training sets, where 20\%,
50\%, or 80\% of the labels are replaced by random noise, respectively. All the
test labels are kept intact for evaluation. Dropout
\citep{srivastava2014dropout} is considered the state-of-the-art method for
learning with corrupted labels \citep{arpit2017closer}. Thus, we compare in
these experiments \mixup{}, dropout, \mixup{} + dropout, and ERM. For \mixup{}, we choose $\alpha
\in \{1, 2, 8, 32\}$; for dropout, we add one dropout layer in each PreAct
block after the ReLU activation layer between two convolution layers, as
suggested in \citep{Zagoruyko2016WRN}. We choose the dropout probability $p \in
\{0.5, 0.7, 0.8, 0.9\}$. For the combination of \mixup{} and dropout, we choose $\alpha\in\{1, 2, 4, 8\}$ and $p\in\{0.3, 0.5, 0.7\}$. These experiments use the PreAct ResNet-18
\citep{he2016identity} model implemented in \citep{cifar-pytorch}. All the
other settings are the same as in Section \ref{sec:cifar}.

We summarize our results in Table \ref{table:corrupt}, where we note the best
test error achieved during the training session, as well as the final test
error after 200 epochs. To quantify the amount of memorization, we also
evaluate the training errors at the last epoch on real labels and corrupted
labels. As the training progresses with a smaller learning rate (e.g. less than
0.01), the ERM model starts to overfit the corrupted labels. When using a large
probability (e.g. 0.7 or 0.8), dropout can effectively reduce overfitting.
\mixup{} with a large $\alpha$ (e.g. 8 or 32) outperforms
dropout on both the best and last epoch test errors, and achieves lower
training error on real labels while remaining resistant to noisy labels. Interestingly, \mixup{} + dropout performs the best of all, showing that the two methods are compatible.

\begin{table}
  \centering
  \begin{tabular}[b]{ll rr rr}
    \toprule
    \multirow{2}{*}{Label corruption} & \multirow{2}{*}{Method} & \multicolumn{2}{c}{Test error} & \multicolumn{2}{c}{Training error} \\
    \cmidrule(lr){3-4} \cmidrule(lr){5-6}
    \rule{0pt}{2ex} & & Best & Last & Real & Corrupted \\
    \midrule
    \multirow{3}{*}{20\%} & ERM & $12.7$ & $16.6$ & $0.05$ & $0.28$ \\
    & ERM + dropout ($p=0.7$) & $8.8$ & $10.4$ & $5.26$ & $83.55$ \\
     & \mixup{} ($\alpha=8$)& $\bf 5.9$ & $6.4$ & $2.27$ & $86.32$ \\
    & \mixup{} + dropout ($\alpha=4, p=0.1$) & $6.2$ & $\bf 6.2$ & $1.92$ & $85.02$ \\
    \midrule
    \multirow{3}{*}{50\%} & ERM & $18.8$ & $44.6$ & $0.26$ & $0.64$ \\
    & ERM + dropout ($p=0.8$)& $14.1$ & $15.5$ & $12.71$ & $86.98$ \\
    & \mixup{} ($\alpha=32$) & $11.3$ & $12.7$ & $5.84$ & $85.71$ \\
    & \mixup{} + dropout ($\alpha=8, p=0.3$) & $\bf 10.9$ & $\bf 10.9$ & $7.56$ & $87.90$ \\
    \midrule
    \multirow{4}{*}{80\%} & ERM & $36.5$ & $73.9$ & $0.62$ & $0.83$ \\
    & ERM + dropout ($p=0.8$)& $30.9$ & $35.1$ & $29.84$ & $86.37$ \\
    & \mixup{} ($\alpha=32$) & $25.3$ & $30.9$ & $18.92$ & $85.44$ \\
    & \mixup{} + dropout ($\alpha=8, p=0.3$) & $\bf 24.0$ & $\bf 24.8$ & $19.70$ & $87.67$ \\
    \bottomrule
  \end{tabular}
  \caption{Results on the corrupted label experiments for the best models.}
  \label{table:corrupt}
\end{table}

\subsection{Robustness to Adversarial examples}
\label{sec:adversarial}

One undesirable consequence of models trained using ERM is their fragility to
adversarial examples~\citep{SzegedyZSBEGF13}. Adversarial examples are obtained
by adding tiny (visually imperceptible) perturbations to legitimate examples in
order to deteriorate the performance of the model.  The adversarial noise is
generated by ascending the gradient of the loss surface with respect to the
legitimate example. Improving the robustness to adversarial examples is a topic
of active research.

Among the several methods aiming to solve this problem,
some have proposed to penalize the norm of the Jacobian of the model to control
its Lipschitz constant~\citep{drucker1992improving, cisse2017parseval, bartlett2017spectrally,
hein2017formal}. Other approaches perform data augmentation by producing and
training on adversarial examples \citep{goodfellow2014explaining}.
Unfortunately, all of these methods add significant computational overhead to
ERM. Here, we show that
\mixup{} can significantly improve the robustness of neural networks without
hindering the speed of ERM by penalizing the norm of the gradient of the loss w.r.t a given input along the most plausible directions
(e.g. the directions to other training points). Indeed, Figure~\ref{fig:cifar10_interp} shows that \mixup{} results in models having a smaller loss and gradient norm between
examples compared to vanilla ERM.

To assess the robustness of \mixup{} models to adversarial examples, we use
three ResNet-101 models: two of them trained using ERM on ImageNet-2012, and
the third trained using \mixup{}. In the first set of experiments, we study the
robustness of one ERM model and the \mixup{} model against white box attacks.
That is, for each of the two models, we use the model itself to generate
adversarial examples, either using the Fast Gradient Sign Method (FGSM) or the
Iterative FGSM (I-FGSM) methods \citep{goodfellow2014explaining}, allowing a
maximum perturbation of $\epsilon=4$ for every pixel. For I-FGSM, we use 10
iterations with equal step size. In the second set of experiments, we evaluate
robustness against black box attacks. That is, we use the first ERM model to
produce adversarial examples using FGSM and I-FGSM. Then, we test the
robustness of the second ERM model and the \mixup{} model to these examples.
The results of both settings are summarized in Table~\ref{table:adversarial}.

For the FGSM white box attack, the \mixup{} model is $2.7$ times more robust
than the ERM model in terms of Top-1 error. For the FGSM black box attack, the
\mixup{} model is $1.25$ times more robust than the ERM model in terms of Top-1
error. Also, while both \mixup{} and ERM are not robust to white box I-FGSM
attacks, \mixup{} is about $40\%$ more robust than ERM in the black box I-FGSM
setting. Overall, \mixup{} produces neural networks that are significantly more
robust than ERM against adversarial examples in white box and black settings without additional overhead compared to ERM.

\begin{table}
  \begin{subfigure}{0.5\textwidth}
  \begin{center}
  \begin{tabular}[b]{ ll rr rr}
    \toprule
    Metric & Method & FGSM &  I-FGSM\\
    \midrule
    \multirow{2}{*}{Top-1} & ERM    & $    90.7$ & $99.9$\\
                           & \mixup & $\bf 75.2$ & $99.6$\\
    \midrule
    \multirow{2}{*}{Top-5} & ERM    &     $63.1$ & $93.4$\\
                           & \mixup & $\bf 49.1$ & $95.8$\\
    \bottomrule
  \end{tabular}
  \end{center}
  \caption{White box attacks.}
  \end{subfigure}
  \hfill
  \begin{subfigure}{0.5\textwidth}
  \begin{center}
  \begin{tabular}[b]{ ll rr}
    \toprule
    Metric & Method & FGSM &  I-FGSM \\
    \midrule
    \multirow{2}{*}{Top-1} & ERM    &     $57.0$ &     $57.3$\\
                           & \mixup & $\bf 46.0$ & $\bf 40.9$\\
    \midrule
    \multirow{2}{*}{Top-5} & ERM    &     $24.8$ &     $18.1$\\
                           & \mixup & $\bf 17.4$ & $\bf 11.8$\\
    \bottomrule
  \end{tabular}
  \end{center}
  \caption{Black box attacks.}
  \end{subfigure}
  \caption{Classification errors of ERM and \mixup{} models when tested on
  adversarial examples.}
  \label{table:adversarial}
\end{table}

\subsection{Tabular data}
\label{sec:uci}
To further explore the performance of \mixup{} on non-image data, we performed
a series of experiments on six arbitrary classification problems drawn from the
UCI dataset \citep{uci}. The neural networks in this section are
fully-connected, and have two hidden layers of 128 ReLU units. The parameters
of these neural networks are learned using Adam \citep{kingma2014adam}
with default hyper-parameters, over 10 epochs of mini-batches of size 16.
Table~\ref{table:uci} shows that \mixup{} improves the average test error on
four out of the six considered datasets, and never underperforms
ERM.

\begin{table}
  \begin{center}
  \begin{tabular}{ l r r}
  \toprule
  Dataset & ERM &  \mixup{} \\
  \midrule
  Abalone    & $74.0$  & $   73.6$ \\
  Arcene     & $57.6$  & $\bf 48.0$ \\
  Arrhythmia & $56.6$ & $\bf  46.3$\\
  \bottomrule
  \end{tabular}
  \hspace{0.8cm}
  \begin{tabular}{ l r r}
      \toprule
      Dataset & ERM &  \mixup{} \\
      \midrule
      Htru2    & $ 2.0$  & $    2.0$ \\
      Iris     & $21.3$ & $\bf 17.3$ \\
      Phishing & $16.3$ & $    15.2$ \\
      \bottomrule
  \end{tabular}
  \end{center}
  \caption{ERM and \mixup{} classification errors on the UCI datasets.}
  \label{table:uci}
\end{table}

\subsection{Stabilization of Generative Adversarial Networks (GANs)}
\label{sec:gans}

\begin{figure}
    \begin{center}
    \resizebox{\textwidth}{!}{
    \begin{tabular}{c c}
        \toprule
        ERM GAN & \mixup{} GAN ($\alpha = 0.2$)\\
        \midrule
        \includegraphics[width=0.09\textwidth]{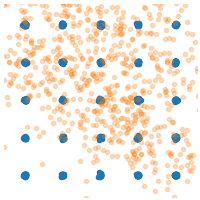}
        \includegraphics[width=0.09\textwidth]{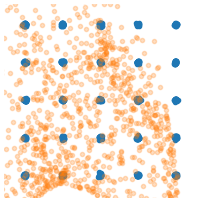}
        \includegraphics[width=0.09\textwidth]{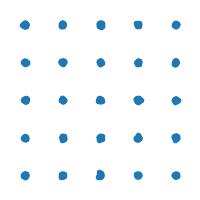}
        \includegraphics[width=0.09\textwidth]{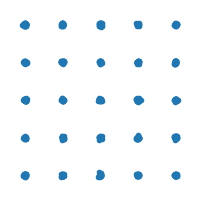}
        \includegraphics[width=0.09\textwidth]{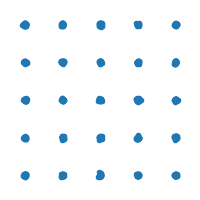} &
        \includegraphics[width=0.09\textwidth]{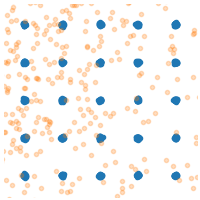}
        \includegraphics[width=0.09\textwidth]{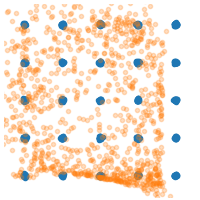}
        \includegraphics[width=0.09\textwidth]{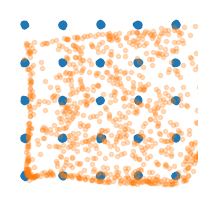}
        \includegraphics[width=0.09\textwidth]{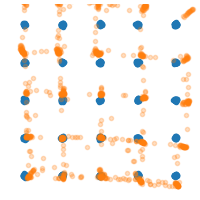}
        \includegraphics[width=0.09\textwidth]{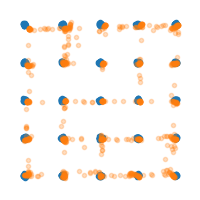}\\
        \includegraphics[width=0.09\textwidth]{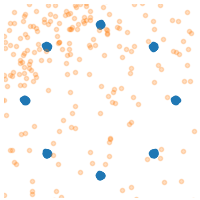}
        \includegraphics[width=0.09\textwidth]{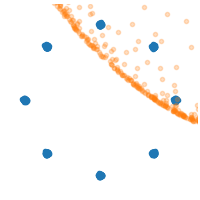}
        \includegraphics[width=0.09\textwidth]{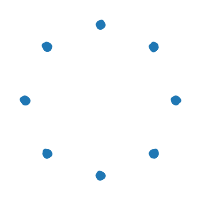}
        \includegraphics[width=0.09\textwidth]{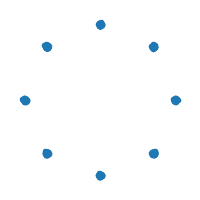}
        \includegraphics[width=0.09\textwidth]{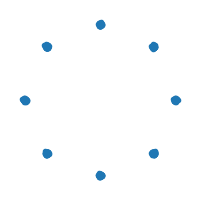}&
        \includegraphics[width=0.09\textwidth]{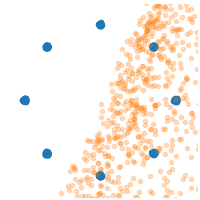}
        \includegraphics[width=0.09\textwidth]{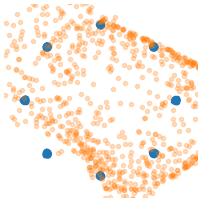}
        \includegraphics[width=0.09\textwidth]{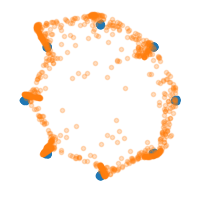}
        \includegraphics[width=0.09\textwidth]{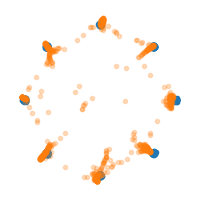}
        \includegraphics[width=0.09\textwidth]{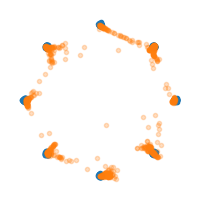}\\
        \bottomrule
    \end{tabular}
    }
    \end{center}
    \caption{Effect of \mixup{} on stabilizing GAN training at iterations 10, 100, 1000, 10000, and 20000.}
    \label{fig:gans}
\end{figure}

Generative Adversarial Networks, also known as GANs
\citep{goodfellow2014generative}, are a powerful family of implicit generative
models.  In GANs, a generator and a discriminator compete against each other to
model a distribution $P$. On the one hand, the generator $g$ competes to
transform noise vectors $z \sim Q$ into fake samples $g(z)$ that resemble real
samples $x \sim P$.  On the other hand, the discriminator competes to
distinguish between real samples $x$ and fake samples $g(z)$. Mathematically,
training a GAN is equivalent to solving the optimization problem
\begin{equation*}
  \max_g \min_d \E_{x, z} \, \ell(d(x), 1) + \ell(d(g(z)), 0),
\end{equation*}
where $\ell$ is the binary cross entropy loss. Unfortunately, solving the
previous min-max equation is a notoriously difficult optimization problem
\citep{goodfellow2016nips}, since the discriminator often provides the
generator with vanishing gradients.  We argue that \mixup{} should stabilize
GAN training because it acts as a regularizer on the gradients of the
discriminator, akin to the binary classifier in Figure~\ref{fig:mixup:toy}.
Then, the smoothness of the discriminator guarantees a stable source of
gradient information to the generator. The \mixup{} formulation of GANs is:
\begin{equation*}
  \max_g \min_d \E_{x, z, \lambda} \, \ell(d(\lambda x + (1 - \lambda) g(z)),
  \lambda).
\end{equation*}
Figure~\ref{fig:gans} illustrates the stabilizing effect of \mixup{} the
training of GAN (orange samples) when modeling two toy datasets (blue samples).
The neural networks in these experiments are fully-connected and have three
hidden layers of 512 ReLU units. The generator network accepts two-dimensional
Gaussian noise vectors. The networks are trained for 20,000 mini-batches of
size 128 using the Adam optimizer with default parameters, where the
discriminator is trained for five iterations before every generator iteration.
The training of \mixup{} GANs seems promisingly robust to hyper-parameter and
architectural choices.

\subsection{Ablation Studies}
\label{sec:ablation}
\mixup{} is a data augmentation method that consists of only two parts: random convex combination of raw inputs, and correspondingly, convex combination of one-hot label encodings. However, there are several design choices to make. For example, on how to augment the inputs, we could have chosen to interpolate the latent representations (i.e. feature maps) of a neural network, and we could have chosen to interpolate only between the nearest neighbors, or only between inputs of the same class. When the inputs to interpolate come from two different classes, we could have chosen to assign a single label to the synthetic input, for example using the label of the input that weights more in the convex combination. To compare \mixup{} with these alternative possibilities, we run a set of ablation study experiments using the PreAct ResNet-18 architecture on the CIFAR-10 dataset.

Specifically, for each of the data augmentation methods, we test two weight decay settings ($10^{-4}$ which works well for \mixup{}, and $5\times 10^{-4}$ which works well for ERM). All the other settings and hyperparameters are the same as reported in Section \ref{sec:cifar}.

\begin{table}
	\centering
	\begin{tabular}[b]{ll rr rr}
		\toprule
		\multirow{2}{*}{Method} & \multirow{2}{*}{Specification} & \multicolumn{2}{c}{Modified} & \multicolumn{2}{c}{Weight decay} \\
		\cmidrule(lr){3-4} \cmidrule(lr){5-6}
		\rule{0pt}{2ex} & & Input & Target & $10^{-4}$ & $5\times 10^{-4}$ \\
		\midrule
		ERM & & \xmark & \xmark & $5.53$ & $5.18$ \\
		\midrule
		\mixup{} & AC + RP & \cmark & \cmark &  $\bf 4.24$ & $4.68$ \\
		& AC + KNN & \cmark & \cmark & $4.98$ & $5.26$ \\
		\midrule
		mix labels and latent & Layer 1 & \cmark & \cmark & $4.44$ & $\bf 4.51$ \\
		representations & Layer 2 & \cmark & \cmark & $4.56$ & $4.61$ \\
		(AC + RP) & Layer 3 & \cmark & \cmark & $5.39$ & $5.55$ \\
		& Layer 4 & \cmark & \cmark & $5.95$ & $5.43$ \\
		& Layer 5 & \cmark & \cmark & $5.39$ & $5.15$ \\
		\midrule
		mix inputs only & SC + KNN \citep{chawla2002smote} & \cmark & \xmark & $5.45$ & $5.52$ \\
		& AC + KNN & \cmark & \xmark & $5.43$ & $5.48$ \\
		& SC + RP & \cmark & \xmark & $5.23$ & $5.55$ \\
		& AC + RP & \cmark & \xmark & $5.17$ & $5.72$ \\
		\midrule
		label smoothing & $\epsilon=0.05$ & \xmark & \cmark & $5.25$ & $5.02$ \\
		\citep{szegedy2016rethinking} & $\epsilon=0.1$ & \xmark & \cmark & $5.33$ & $5.17$ \\
		& $\epsilon=0.2$ & \xmark & \cmark & $5.34$ & $5.06$ \\
		\midrule
		mix inputs + & $\epsilon=0.05$ & \cmark & \cmark & $5.02$ & $5.40$ \\
		label smoothing & $\epsilon=0.1$ & \cmark & \cmark & $5.08$ & $5.09$ \\
		(AC + RP) & $\epsilon=0.2$ & \cmark & \cmark & $4.98$ & $5.06$ \\
		& $\epsilon=0.4$ & \cmark & \cmark & $5.25$ & $5.39$ \\
		\midrule
		add Gaussian noise & $\sigma=0.05$ & \cmark & \xmark & $5.53$ & $5.04$ \\
		to inputs & $\sigma=0.1$ & \cmark & \xmark & $6.41$ & $5.86$ \\
		& $\sigma=0.2$ & \cmark & \xmark & $7.16$ & $7.24$ \\
		\bottomrule
	\end{tabular}
	\caption{Results of the ablation studies on the CIFAR-10 dataset. Reported are the median test errors of the last 10 epochs. A tick (\cmark) means the component is different from standard ERM training, whereas a cross (\xmark) means it follows the standard training practice. AC: mix between all classes. SC: mix within the same class. RP: mix between random pairs. KNN: mix between k-nearest neighbors (k=200). Please refer to the text for details about the experiments and interpretations.}
	\label{table:ablation}
\end{table}

To compare interpolating raw inputs with interpolating latent representations, we test on random convex combination of the learned representations before each residual block (denoted Layer 1-4) or before the uppermost ``average pooling + fully connected" layer (denoted Layer 5). To compare mixing random pairs of inputs (RP) with mixing nearest neighbors (KNN), we first compute the 200 nearest neighbors for each training sample, either from the same class (SC) or from all the classes (AC). Then during training, for each sample in a minibatch, we replace the sample with a synthetic sample by convex combination with a random draw from its nearest neighbors. To compare mixing all the classes (AC) with mixing within the same class (SC), we convex combine a minibatch with a random permutation of its sample index, where the permutation is done in a per-batch basis (AC) or a per-class basis (SC). To compare mixing inputs and labels with mixing inputs only, we either use a convex combination of the two one-hot encodings as the target, or select the one-hot encoding of the closer training sample as the target. For label smoothing, we follow \cite{szegedy2016rethinking} and use $\frac{\epsilon}{10}$ as the target for incorrect classes, and $1-\frac{9\epsilon}{10}$ as the target for the correct class.　Adding Gaussian noise to inputs is used as another baseline. We report the median test errors of the last 10 epochs. Results are shown in Table \ref{table:ablation}.

From the ablation study experiments, we have the following observations. First, \mixup{} is the best data augmentation method we test, and is significantly better than the second best method (mix input + label smoothing). Second, the effect of regularization can be seen by comparing the test error with a small weight decay ($10^{-4}$) with a large one ($5\times 10^{-4}$). For example, for ERM a large weight decay works better, whereas for \mixup{} a small weight decay is preferred, confirming its regularization effects. We also see an increasing advantage of large weight decay when interpolating in higher layers of latent representations, indicating decreasing strength of regularization.
Among all the input interpolation methods, mixing random pairs from all classes (AC + RP) has the strongest regularization effect. Label smoothing and adding Gaussian noise have a relatively small regularization effect. Finally, we note that the SMOTE algorithm \citep{chawla2002smote} does not lead to a noticeable gain in performance.

	\section{Related Work}
\label{sec:related}

Data augmentation lies at the heart of all successful applications of deep
learning, ranging from image classification~\citep{krizhevsky2012imagenet} to
speech recognition~\citep{graves2013speech, amodei2016deep}. In all cases,
substantial domain knowledge is leveraged to design suitable data
transformations leading to improved generalization. In image classification,
for example, one routinely uses rotation, translation, cropping, resizing,
flipping~\citep{lecun98, simonyan2014very}, and random
erasing~\citep{zhong2017random} to enforce visually plausible invariances in
the model through the training data. Similarly, in speech recognition, noise
injection is a prevalent practice to improve the robustness and accuracy of the
trained models~\citep{amodei2016deep}. 

More related to \mixup{}, \citet{chawla2002smote} propose to augment the rare class in an imbalanced dataset by interpolating the nearest neighbors; \citet{devries2017dataset} show that interpolation
and extrapolation the nearest neighbors of the same class in feature space can improve generalization. However, their
proposals only operate among the nearest neighbors within a certain class at the input / feature level, and hence does not account for changes in
the corresponding labels. Recent approaches have also proposed to regularize
the output distribution of a neural network by label
smoothing~\citep{szegedy2016rethinking}, or penalizing high-confidence softmax
distributions~\citep{pereyra2017regularizing}. These methods bear similarities
with \mixup{} in the sense that supervision depends on multiple smooth labels,
rather than on single hard labels as in traditional ERM. However, the label
smoothing in these works is applied or regularized independently from the
associated feature values.

\mixup{} enjoys several desirable aspects of previous data augmentation and
regularization schemes without suffering from their drawbacks. Like the method
of~\cite{devries2017dataset}, it does not require significant domain knowledge.
Like label smoothing, the supervision of every example is not overly dominated
by the ground-truth label. Unlike both of these approaches, the \mixup{}
transformation establishes a linear relationship between data augmentation and
the supervision signal. We believe that this leads to a strong regularizer that
improves generalization as demonstrated by our experiments. The linearity
constraint, through its effect on the derivatives of the function approximated,
also relates \mixup{} to other methods such as Sobolev training of neural
networks~\citep{czarnecki2017sobolev} or WGAN-GP~\citep{gulrajani2017improved}.

    \section{Discussion}
\label{sec:discussion}

We have proposed \mixup{}, a data-agnostic and straightforward data
augmentation principle.  We have shown that \mixup{} is a form of vicinal risk
minimization, which trains on virtual examples constructed as the linear
interpolation of two random examples from the training set and their labels.
Incorporating \mixup{} into existing training pipelines reduces to a few lines
of code, and introduces little or no computational overhead.  Throughout an
extensive evaluation, we have shown that \mixup{} improves the generalization
error of state-of-the-art models on ImageNet, CIFAR, speech, and tabular
datasets. Furthermore, \mixup{} helps to combat memorization of corrupt labels,
sensitivity to adversarial examples, and instability in adversarial training.

In our experiments, the following trend is consistent: with increasingly large
$\alpha$, the training error on real data increases, while the generalization
gap decreases. This sustains our hypothesis that \mixup{} implicitly controls
model complexity. However, we do not yet have a good theory for understanding
the `sweet spot' of this bias-variance trade-off. For example, in CIFAR-10
classification we can get very low training error on real data even when
$\alpha \to \infty$ (i.e., training \emph{only} on averages of pairs of real
examples), whereas in ImageNet classification, the training error on real data
increases significantly with $\alpha \to \infty$. Based on our ImageNet and
Google commands experiments with different model architectures, we conjecture
that increasing the model capacity would make training error less sensitive to
large $\alpha$, hence giving \mixup{} a more significant advantage. 

\mixup{} also opens up several possibilities for further exploration. First, is
it possible to make similar ideas work on other types of supervised learning
problems, such as regression and structured prediction? While generalizing
\mixup{} to regression problems is straightforward, its application to
structured prediction problems such as image segmentation remains less obvious.
Second, can similar methods prove helpful beyond supervised learning? The
interpolation principle seems like a reasonable inductive bias which might also
help in unsupervised, semi-supervised, and reinforcement learning. Can we
extend \mixup{} to feature-label extrapolation to guarantee a robust model
behavior far away from the training data? Although our discussion of these
directions is still speculative, we are excited about the possibilities
\mixup{} opens up, and hope that our observations will prove useful for future
development. 

\ificlrfinal
\section*{Acknowledgements}
We would like to thank Priya Goyal, Yossi Adi and the PyTorch team. We also thank the Anonymous Review 2 for proposing the \mixup{} + dropout experiments.
\fi

    {\small
        \bibliography{mixup}

\begin{thebibliography}{45}
\providecommand{\natexlab}[1]{#1}
\providecommand{\url}[1]{\texttt{#1}}
\expandafter\ifx\csname urlstyle\endcsname\relax
  \providecommand{\doi}[1]{doi: #1}\else
  \providecommand{\doi}{doi: \begingroup \urlstyle{rm}\Url}\fi

\bibitem[Amodei et~al.(2016)Amodei, Ananthanarayanan, Anubhai, Bai, Battenberg,
  Case, Casper, Catanzaro, Cheng, Chen, et~al.]{amodei2016deep}
D.~Amodei, S.~Ananthanarayanan, R.~Anubhai, J.~Bai, E.~Battenberg, C.~Case,
  J.~Casper, B.~Catanzaro, Q.~Cheng, G.~Chen, et~al.
\newblock Deep speech 2: End-to-end speech recognition in {E}nglish and
  {M}andarin.
\newblock In \emph{ICML}, 2016.

\bibitem[Arpit et~al.(2017)Arpit, Jastrzebski, Ballas, Krueger, Bengio, Kanwal,
  Maharaj, Fischer, Courville, Bengio, et~al.]{arpit2017closer}
D.~Arpit, S.~Jastrzebski, N.~Ballas, D.~Krueger, E.~Bengio, M.~S. Kanwal,
  T.~Maharaj, A.~Fischer, A.~Courville, Y.~Bengio, et~al.
\newblock A closer look at memorization in deep networks.
\newblock \emph{ICML}, 2017.

\bibitem[Bartlett et~al.(2017)Bartlett, Foster, and
  Telgarsky]{bartlett2017spectrally}
P.~Bartlett, D.~J. Foster, and M.~Telgarsky.
\newblock Spectrally-normalized margin bounds for neural networks.
\newblock \emph{NIPS}, 2017.

\bibitem[Chapelle et~al.(2000)Chapelle, Weston, Bottou, and Vapnik]{vicinal}
O.~Chapelle, J.~Weston, L.~Bottou, and V.~Vapnik.
\newblock Vicinal risk minimization.
\newblock \emph{NIPS}, 2000.

\bibitem[Chawla et~al.(2002)Chawla, Bowyer, Hall, and
  Kegelmeyer]{chawla2002smote}
N.~V. Chawla, K.~W. Bowyer, L.~O. Hall, and W.~P. Kegelmeyer.
\newblock {SMOTE}: synthetic minority over-sampling technique.
\newblock \emph{Journal of artificial intelligence research}, 16:\penalty0
  321--357, 2002.

\bibitem[Chelba et~al.(2013)Chelba, Mikolov, Schuster, Ge, Brants, Koehn, and
  Robinson]{chelba2013one}
C.~Chelba, T.~Mikolov, M.~Schuster, Q.~Ge, T.~Brants, P.~Koehn, and
  T.~Robinson.
\newblock One billion word benchmark for measuring progress in statistical
  language modeling.
\newblock \emph{arXiv}, 2013.

\bibitem[Cisse et~al.(2017)Cisse, Bojanowski, Grave, Dauphin, and
  Usunier]{cisse2017parseval}
M.~Cisse, P.~Bojanowski, E.~Grave, Y.~Dauphin, and N.~Usunier.
\newblock Parseval networks: Improving robustness to adversarial examples.
\newblock \emph{ICML}, 2017.

\bibitem[Czarnecki et~al.(2017)Czarnecki, Osindero, Jaderberg, {\'S}wirszcz,
  and Pascanu]{czarnecki2017sobolev}
W.~M. Czarnecki, S.~Osindero, M.~Jaderberg, G.~{\'S}wirszcz, and R.~Pascanu.
\newblock Sobolev training for neural networks.
\newblock \emph{NIPS}, 2017.

\bibitem[DeVries \& Taylor(2017)DeVries and Taylor]{devries2017dataset}
T.~DeVries and G.~W. Taylor.
\newblock Dataset augmentation in feature space.
\newblock \emph{ICLR Workshops}, 2017.

\bibitem[Drucker \& Le~Cun(1992)Drucker and Le~Cun]{drucker1992improving}
H.~Drucker and Y.~Le~Cun.
\newblock Improving generalization performance using double backpropagation.
\newblock \emph{IEEE Transactions on Neural Networks}, 3\penalty0 (6):\penalty0
  991--997, 1992.

\bibitem[Goodfellow(2016)]{goodfellow2016nips}
I.~Goodfellow.
\newblock Tutorial: Generative adversarial networks.
\newblock \emph{NIPS}, 2016.

\bibitem[Goodfellow et~al.(2014)Goodfellow, Pouget-Abadie, Mirza, Xu,
  Warde-Farley, Ozair, Courville, and Bengio]{goodfellow2014generative}
I.~Goodfellow, J.~Pouget-Abadie, M.~Mirza, B.~Xu, D.~Warde-Farley, S.~Ozair,
  A.~Courville, and Y.~Bengio.
\newblock Generative adversarial nets.
\newblock \emph{NIPS}, 2014.

\bibitem[Goodfellow et~al.(2015)Goodfellow, Shlens, and
  Szegedy]{goodfellow2014explaining}
I.~J. Goodfellow, J.~Shlens, and C.~Szegedy.
\newblock Explaining and harnessing adversarial examples.
\newblock \emph{ICLR}, 2015.

\bibitem[Goyal et~al.(2017)Goyal, Doll{\'a}r, Girshick, Noordhuis, Wesolowski,
  Kyrola, Tulloch, Jia, and He]{goyal2017accurate}
P.~Goyal, P.~Doll{\'a}r, R.~Girshick, P.~Noordhuis, L.~Wesolowski, A.~Kyrola,
  A.~Tulloch, Y.~Jia, and K.~He.
\newblock Accurate, large minibatch {SGD}: Training {I}mage{Net} in 1 hour.
\newblock \emph{arXiv}, 2017.

\bibitem[Graves et~al.(2013)Graves, Mohamed, and Hinton]{graves2013speech}
A.~Graves, A.-r. Mohamed, and G.~Hinton.
\newblock Speech recognition with deep recurrent neural networks.
\newblock In \emph{ICASSP}. IEEE, 2013.

\bibitem[Gulrajani et~al.(2017)Gulrajani, Ahmed, Arjovsky, Dumoulin, and
  Courville]{gulrajani2017improved}
I.~Gulrajani, F.~Ahmed, M.~Arjovsky, V.~Dumoulin, and A.~Courville.
\newblock Improved training of {W}asserstein {GAN}s.
\newblock \emph{NIPS}, 2017.

\bibitem[Harvey et~al.(2017)Harvey, Liaw, and Mehrabian]{harvey2017nearly}
N.~Harvey, C.~Liaw, and A.~Mehrabian.
\newblock Nearly-tight {VC}-dimension bounds for piecewise linear neural
  networks.
\newblock \emph{JMLR}, 2017.

\bibitem[He et~al.(2016)He, Zhang, Ren, and Sun]{he2016identity}
K.~He, X.~Zhang, S.~Ren, and J.~Sun.
\newblock Identity mappings in deep residual networks.
\newblock \emph{ECCV}, 2016.

\bibitem[Hein \& Andriushchenko(2017)Hein and Andriushchenko]{hein2017formal}
M.~Hein and M.~Andriushchenko.
\newblock Formal guarantees on the robustness of a classifier against
  adversarial manipulation.
\newblock \emph{NIPS}, 2017.

\bibitem[Hinton et~al.(2012)Hinton, Deng, Yu, Dahl, Mohamed, Jaitly, Senior,
  Vanhoucke, Nguyen, Sainath, et~al.]{hinton2012deep}
G.~Hinton, L.~Deng, D.~Yu, G.~E. Dahl, A.-r. Mohamed, N.~Jaitly, A.~Senior,
  V.~Vanhoucke, P.~Nguyen, T.~N. Sainath, et~al.
\newblock Deep neural networks for acoustic modeling in speech recognition: The
  shared views of four research groups.
\newblock \emph{IEEE Signal Processing Magazine}, 2012.

\bibitem[Huang et~al.(2017)Huang, Liu, van~der Maaten, and
  Weinberger]{huang2017densely}
G.~Huang, Z.~Liu, L.~van~der Maaten, and K.~Q. Weinberger.
\newblock Densely connected convolutional networks.
\newblock \emph{CVPR}, 2017.

\bibitem[Kingma \& Ba(2015)Kingma and Ba]{kingma2014adam}
D.~Kingma and J.~Ba.
\newblock Adam: A method for stochastic optimization.
\newblock \emph{ICLR}, 2015.

\bibitem[Krizhevsky et~al.(2012)Krizhevsky, Sutskever, and
  Hinton]{krizhevsky2012imagenet}
A.~Krizhevsky, I.~Sutskever, and G.~E. Hinton.
\newblock Image{Net} classification with deep convolutional neural networks.
\newblock \emph{NIPS}, 2012.

\bibitem[Lecun et~al.(2001)Lecun, Bottou, Bengio, and Haffner]{lecun98}
Y.~Lecun, L.~Bottou, Y.~Bengio, and P.~Haffner.
\newblock Gradient-based learning applied to document recognition.
\newblock \emph{Proceedings of IEEE}, 2001.

\bibitem[Lichman(2013)]{uci}
M.~Lichman.
\newblock {UCI} machine learning repository, 2013.

\bibitem[Liu(2017)]{cifar-pytorch}
K.~Liu, 2017.
\newblock URL \url{https://github.com/kuangliu/pytorch-cifar}.

\bibitem[Pereyra et~al.(2017)Pereyra, Tucker, Chorowski, Kaiser, and
  Hinton]{pereyra2017regularizing}
G.~Pereyra, G.~Tucker, J.~Chorowski, {\L}.~Kaiser, and G.~Hinton.
\newblock Regularizing neural networks by penalizing confident output
  distributions.
\newblock \emph{ICLR Workshops}, 2017.

\bibitem[Russakovsky et~al.(2015)Russakovsky, Deng, Su, Krause, Satheesh, Ma,
  Huang, Karpathy, Khosla, Bernstein, Berg, and Fei-Fei]{ILSVRC15}
O.~Russakovsky, J.~Deng, H.~Su, J.~Krause, S.~Satheesh, S.~Ma, Z.~Huang,
  A.~Karpathy, A.~Khosla, M.~Bernstein, A.~C. Berg, and L.~Fei-Fei.
\newblock Image{Net} large scale visual recognition challenge.
\newblock \emph{IJCV}, 2015.

\bibitem[Silver et~al.(2016)Silver, Huang, Maddison, Guez, Sifre, Van
  Den~Driessche, Schrittwieser, Antonoglou, Panneershelvam, Lanctot,
  et~al.]{silver2016mastering}
D.~Silver, A.~Huang, C.~J. Maddison, A.~Guez, L.~Sifre, G.~Van Den~Driessche,
  J.~Schrittwieser, I.~Antonoglou, V.~Panneershelvam, M.~Lanctot, et~al.
\newblock Mastering the game of {Go} with deep neural networks and tree search.
\newblock \emph{Nature}, 2016.

\bibitem[Simard et~al.(1998)Simard, LeCun, Denker, and
  Victorri]{simard1998transformation}
P.~Simard, Y.~LeCun, J.~Denker, and B.~Victorri.
\newblock Transformation invariance in pattern recognition—tangent distance
  and tangent propagation.
\newblock \emph{Neural networks: tricks of the trade}, 1998.

\bibitem[Simonyan \& Zisserman(2015)Simonyan and Zisserman]{simonyan2014very}
K.~Simonyan and A.~Zisserman.
\newblock Very deep convolutional networks for large-scale image recognition.
\newblock \emph{ICLR}, 2015.

\bibitem[Springenberg et~al.(2015)Springenberg, Dosovitskiy, Brox, and
  Riedmiller]{springenberg2014striving}
J.~T. Springenberg, A.~Dosovitskiy, T.~Brox, and M.~Riedmiller.
\newblock Striving for simplicity: The all convolutional net.
\newblock \emph{ICLR Workshops}, 2015.

\bibitem[Srivastava et~al.(2014)Srivastava, Hinton, Krizhevsky, Sutskever, and
  Salakhutdinov]{srivastava2014dropout}
N.~Srivastava, G.~E. Hinton, A.~Krizhevsky, I.~Sutskever, and R.~Salakhutdinov.
\newblock Dropout: a simple way to prevent neural networks from overfitting.
\newblock \emph{Journal of Machine Learning Research}, 15\penalty0
  (1):\penalty0 1929--1958, 2014.

\bibitem[Szegedy et~al.(2014)Szegedy, Zaremba, Sutskever, Bruna, Erhan,
  Goodfellow, and Fergus]{SzegedyZSBEGF13}
C.~Szegedy, W.~Zaremba, I.~Sutskever, J.~Bruna, D.~Erhan, I.~J. Goodfellow, and
  R.~Fergus.
\newblock Intriguing properties of neural networks.
\newblock \emph{ICLR}, 2014.

\bibitem[Szegedy et~al.(2016)Szegedy, Vanhoucke, Ioffe, Shlens, and
  Wojna]{szegedy2016rethinking}
C.~Szegedy, V.~Vanhoucke, S.~Ioffe, J.~Shlens, and Z.~Wojna.
\newblock Rethinking the {I}nception architecture for computer vision.
\newblock \emph{Proceedings of the IEEE Conference on Computer Vision and
  Pattern Recognition}, 2016.

\bibitem[Vapnik(1998)]{vapnik98}
V.~N. Vapnik.
\newblock \emph{Statistical learning theory}.
\newblock J. Wiley, 1998.

\bibitem[Vapnik \& Chervonenkis(1971)Vapnik and
  Chervonenkis]{vapnik1971uniform}
V.~Vapnik and A.~Y. Chervonenkis.
\newblock On the uniform convergence of relative frequencies of events to their
  probabilities.
\newblock \emph{Theory of Probability and its Applications}, 1971.

\bibitem[Veit(2017)]{dense-andreas}
A.~Veit, 2017.
\newblock URL \url{https://github.com/andreasveit}.

\bibitem[Warden(2017)]{commands}
P.~Warden, 2017.
\newblock URL
  \url{https://research.googleblog.com/2017/08/launching-speech-commands-dataset.html}.

\bibitem[Xie et~al.(2016)Xie, Girshick, Doll{\'a}r, Tu, and
  He]{xie2016aggregated}
S.~Xie, R.~Girshick, P.~Doll{\'a}r, Z.~Tu, and K.~He.
\newblock Aggregated residual transformations for deep neural networks.
\newblock \emph{CVPR}, 2016.

\bibitem[Zagoruyko \& Komodakis(2016{\natexlab{a}})Zagoruyko and
  Komodakis]{Zagoruyko2016WRN}
S.~Zagoruyko and N.~Komodakis.
\newblock Wide residual networks.
\newblock \emph{BMVC}, 2016{\natexlab{a}}.

\bibitem[Zagoruyko \& Komodakis(2016{\natexlab{b}})Zagoruyko and
  Komodakis]{wide-sergey}
S.~Zagoruyko and N.~Komodakis, 2016{\natexlab{b}}.
\newblock URL \url{https://github.com/szagoruyko/wide-residual-networks}.

\bibitem[{Zhang} et~al.(2017){Zhang}, {Bengio}, {Hardt}, {Recht}, and
  {Vinyals}]{2016arXiv161103530Z}
C.~{Zhang}, S.~{Bengio}, M.~{Hardt}, B.~{Recht}, and O.~{Vinyals}.
\newblock {Understanding deep learning requires rethinking generalization}.
\newblock \emph{ICLR}, 2017.

\bibitem[Zhang(2017)]{random-labels}
C.~Zhang, 2017.
\newblock URL \url{https://github.com/pluskid/fitting-random-labels}.

\bibitem[Zhong et~al.(2017)Zhong, Zheng, Kang, Li, and Yang]{zhong2017random}
Z.~Zhong, L.~Zheng, G.~Kang, S.~Li, and Y.~Yang.
\newblock Random erasing data augmentation.
\newblock \emph{arXiv}, 2017.

\end{thebibliography}
        \bibliographystyle{iclr2018}}
\end{document}